\documentclass[acmtog]{acmart}
\acmSubmissionID{916}

\usepackage{booktabs} 
\usepackage{float}

\usepackage[ruled]{algorithm2e} 

\SetAlFnt{\small}
\SetAlCapFnt{\small}
\SetAlCapNameFnt{\small}
\SetAlCapHSkip{0pt}

\setcopyright{rightsretained} 
\acmJournal{TOG}
\acmVolume{43}
\acmNumber{4}
\acmArticle{84}
\acmYear{2024}
\acmMonth{7}
\acmDOI{10.1145/3658217}

\newcommand{\edit}[1]{\textcolor{black}{#1}}
\newcommand{\yanpei}[1]{\textcolor{black}{#1}}
\newcommand{\gmh}[1]{\textcolor{black}{#1}}
\newcommand{\todo}[1]{\textcolor{black}{#1}}
\newcommand{\final}[1]{\textcolor{black}{#1}}

\newcommand*{\method}{CharacterGen}
\newcommand*{\dataset}{Anime3D}

\begin{document}
\title{\method{}: Efficient 3D Character Generation from Single Images with Multi-View Pose Canonicalization}

\author{Hao-Yang Peng}
\email{phy22@mails.tsinghua.edu.cn}
\affiliation{%
 \institution{BNRist, Department of Computer Science and Technology, Tsinghua University}
 \city{Beijing}
 \country{China}}
\author{Jia-Peng Zhang}
\email{zhangjp20@mails.tsinghua.edu.cn}
\affiliation{%
 \institution{Zhili College, Tsinghua University}
 \city{Beijing}
 \country{China}}
\author{Meng-Hao Guo}
\email{gmh20@mails.tsinghua.edu.cn}
\affiliation{%
 \institution{BNRist, Department of Computer Science and Technology, Tsinghua University}
 \city{Beijing}
 \country{China}}
\author{Yan-Pei Cao}
\email{caoyanpei@gmail.com}
\affiliation{%
\institution{VAST}
\city{Beijing}
\country{China}}
\author{Shi-Min Hu}
\email{shimin@tsinghua.edu.cn}
\affiliation{%
\institution{BNRist, Department of Computer Science and Technology, Tsinghua University}
\city{Beijing}
\country{China}}


\begin{abstract}

\yanpei{In the field of digital content creation, generating high-quality 3D characters from single images is challenging, especially given the complexities of various body poses and the issues of self-occlusion and pose ambiguity. In this paper, we present CharacterGen, a framework developed to efficiently generate 3D characters. CharacterGen introduces a streamlined generation pipeline along with an image-conditioned multi-view diffusion model. This model effectively calibrates input poses to a canonical form while retaining key attributes of the input image, thereby addressing the challenges posed by diverse poses. A transformer-based, generalizable sparse-view reconstruction model is the other core component of our approach, facilitating the creation of detailed 3D models from multi-view images. We also adopt a texture-back-projection strategy to produce high-quality texture maps. Additionally, we have curated a dataset of anime characters, rendered in multiple poses and views, to train and evaluate our model. Our approach has been thoroughly evaluated through quantitative and qualitative experiments, showing its proficiency in generating 3D characters with high-quality shapes and textures, ready for downstream applications such as rigging and animation.}
\end{abstract}

\begin{teaserfigure}
  \centering
  \includegraphics[width=0.95\linewidth]{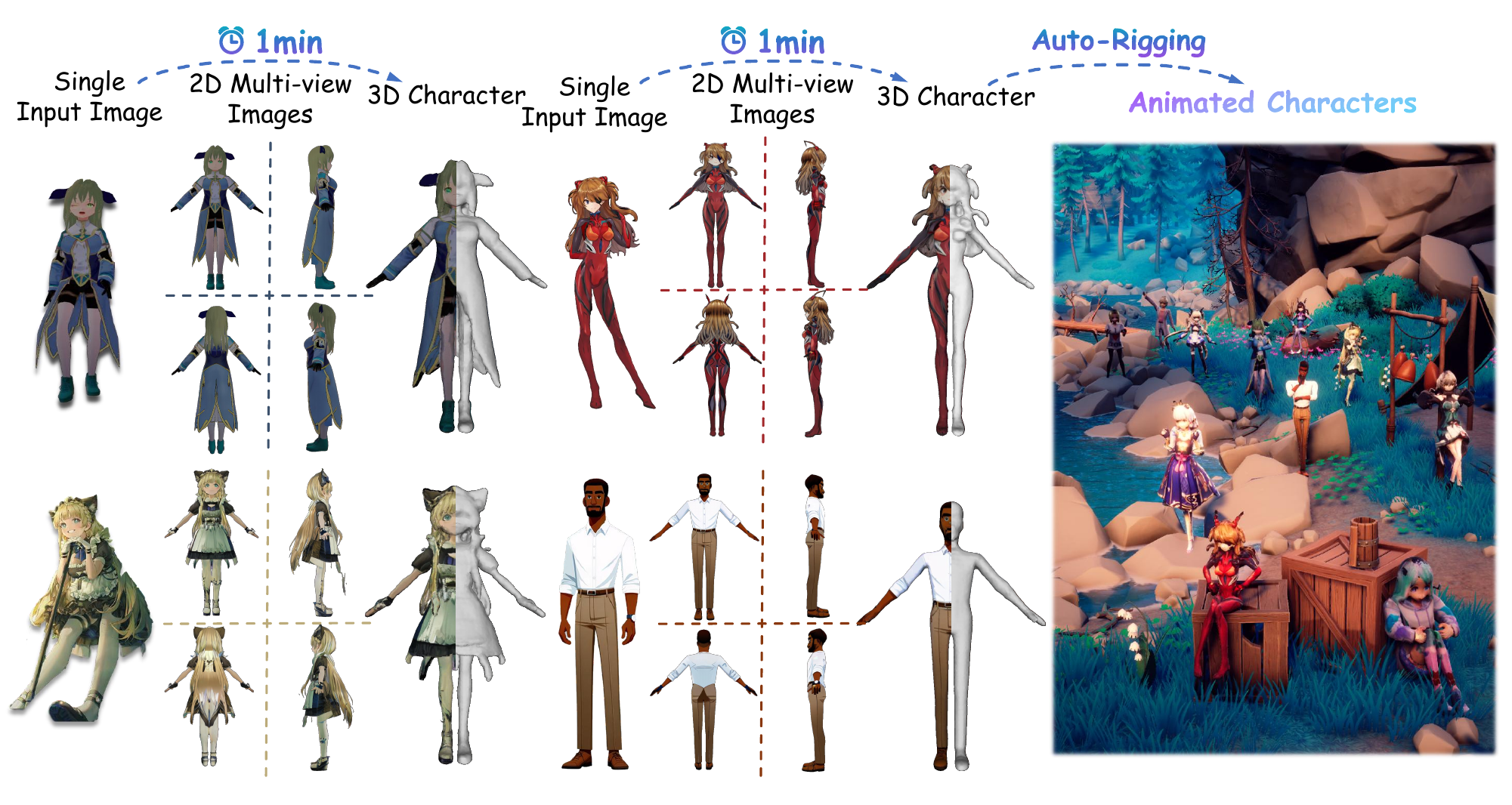}
  \caption{CharacterGen is an efficient 3D character generation framework. It takes a single input image and generates high-quality 3D character mesh in a canonical pose with consistent appearance, suitable for downstream rigging and animation workflows. $\copyright$ kinoko7}
  \label{fig:teaser}
\end{teaserfigure}

%
%

%
%


\maketitle

\section{Introduction}

\yanpei{The digital content industry's rapid evolution has made the creation of high-quality 3D content a pivotal aspect across various domains, including film, video gaming, online streaming, and virtual reality (VR). Although manually modeled 3D content can attain exceptional quality, the significant time and labor investment required presents a substantial bottleneck. Addressing this, there has been a notable influx of exciting research~\cite{tang2023makeit3d, dreamsparse, dreambooth3d, qian2023magic123, wang2023imagedream, nerdi, realfusion, jun2023shap, DBLP:journals/corr/abs-2212-08751, alwala2022pretrain, mcc, Pixel2Mesh, Pixel2Mesh++, one12345, one12345++, wonder3d} focused on generating 3D models from single images. This approach substantially lowers the barrier to entry for novice users, democratizing access to 3D content creation and potentially revolutionizing the field.}

\yanpei{3D character models often feature complex articulations, leading to frequent self-occlusion in 2D images that pose significant challenges in reconstruction, generation, and animation. Moreover, these characters may assume a range of body poses, including some that are rare and challenging to accurately interpret, leading to a diverse yet imbalanced data domain. These complexities hinder the effective generation, rigging, and animating of such models. As a result, general 3D generation techniques~\cite{poole2022dreamfusion, chen2023fantasia3d, wang2023prolificdreamer, lin2023magic3d, wang2023score, latentnerf} and single-view 3D reconstruction methods~\cite{thai20213drecon, hong2023lrm, Pixel2Mesh, Pixel2Mesh++, mcc, one12345, alwala2022pretrain} often fall short in delivering optimal outcomes. Prior research~\cite{cao2023dreamavatar, huang2023dreamwaltz, kolotouros2023dreamhuman, huang2024tech, liao2024tada, xiu2022icon, DBLP:conf/cvpr/XiuYCTB23, saito2019pifu, saito2020pifuhd, pamir, rodin} has explored the use of parametric models of 3D human bodies~\cite{loper2015smpl, pavlakos2019expressive, alldieck2021imghum, flame} as 3D priors. However, these methods are predominantly tailored to realistic human proportions and relatively tight clothing, limiting their applicability. This constraint is especially noticeable in the context of stylized characters, known for their exaggerated body proportions and complex clothing designs, which challenge the adaptability and effectiveness of these approaches.}


\yanpei{In this paper, we introduce \method{}, a new approach for 3D character generation in a canonical pose from a single image. Our method stands out significantly from previous ones by allowing any input body pose in the input image and outputting a clean 3D character model. The foundational principle of \method{} hinges on simultaneously canonicalizing body poses and producing consistent multi-view images during the generation process. This is achieved by transforming each pose into a canonical ``A-pose'', a stance widely utilized in 3D character modeling, while concurrently ensuring image consistency across multiple views. This dual approach effectively addresses the challenges of self-occlusion and ambiguous human poses, significantly streamlining subsequent reconstruction, rigging, and animation stages.}

\yanpei{
Our 3D character generation approach has two tightly interconnected stages: initially, lifting a single image to multiple viewpoints while simultaneously canonicalizing the input pose; following this, we proceed to reconstruct a 3D character using this canonical pose. This method is supported by \emph{two key insights}: \emph{firstly}, it incorporates established principles and successful techniques from recent advancements in controllable image generation~\cite{zhang2023adding, ye2023ip-adapter}; \emph{secondly}, it overcomes the challenges associated with sparse-view reconstruction for 3D characters. By focusing on the canonical pose, in which the geometric and texture structures are more clearly defined and self-occlusion is minimized, our approach simplifies the task of reconstructing both geometry and texture from limited views.
The first stage involves a diffusion-based, image-conditioned multi-view generation model~\cite{liu2023syncdreamer, wang2023imagedream, Zero123++, mvdiffusion}, adept at capturing and translating both the global and local character features from the input image to the canonical pose, which further facilitates the generation of consistent canonical pose images across multiple views. The second stage employs a transformer-based, generalizable sparse-view reconstruction model~\cite{hong2023lrm}. This model is key to generating a coarsely textured 3D character model from the images produced in the first stage. We further refine the model's texture resolution through projective texture mapping and Poisson Blending~\cite{poisson}, achieving a detailed final model. Furthermore, generating characters in a canonical pose also significantly benefits downstream applications, such as rigging and animation.} \final{We show the generated 3D characters with animation in Fig.~\ref{fig:teaser}.} Our whole generation process takes less than 1 minute.

\yanpei{
To train our pipeline, we have compiled a multi-pose, multi-view character dataset, focusing on anime characters due to their widespread availability online, notably on platforms such as VRoid Hub. 
We have amassed a collection of 13,746 characters and rendered these from various viewpoints across multiple body poses. This extensive collection has been organized into a dataset that we refer to as \dataset{}. 
}





\yanpei{
In summary, our paper makes the following key contributions:
\begin{itemize}
\item An image-conditioned diffusion model that effectively generates multi-view consistent images of characters in a controlled canonical pose from varying input poses, addressing challenges such as self-occlusion and pose ambiguity.
\item A streamlined pipeline combining our diffusion model for multi-view image generation and a transformer-based reconstruction model. This pipeline efficiently transforms single-view inputs into detailed 3D character models.
\item A curated dataset of 13,746 anime characters, rendered in multiple poses and views, providing a diverse training and evaluation resource for our model and future research in 3D character generation.
\end{itemize}
}

\section{Related Works}

\edit{This section mainly discusses related works on diffusion-based 3D objects and avatar generation. Our \method{} also adopts transformer-based reconstruction models~\cite{hong2023lrm, DBLP:journals/corr/abs-2311-06214} for efficient 3D character generation. Space limitations preclude discussion of 3D human reconstruction works~\cite{saito2019pifu,saito2020pifuhd,xiu2022icon,DBLP:conf/cvpr/XiuYCTB23,cha2023}.}


\subsection{Diffusion-Based 3D Object Generation}

Recently, diffusion methods have shown a strong ability to guide 3D object generation tasks in the past year. Pioneering work of  DreamFusion~\cite{poole2022dreamfusion} and SJC~\cite{wang2023score} utilize score distillation sampling (SDS) to provide gradient guidance from pre-trained 2D diffusion models for text-to-3D generation tasks. 
Magic3D~\cite{lin2023magic3d} and Fantasia3D~\cite{chen2023fantasia3d} utilize an implicit tetrahedral field to support rendering with high resolution in the refinement stage.
ProlificDreamer~\cite{wang2023prolificdreamer} proposes VSD to distill gradient scores from a LoRA network to better learn the distribution of 3D objects.  
Zero123~\cite{liu2023zero} presents a novel diffusion model to generate multi-view images that conform to the input image with given camera poses. Magic123~\cite{qian2023magic123} combines both SDS and Zero123 guidance for generating 3D objects from image prompts, and adopts reconstruction loss to enhance front-view texture quality.
MVDream~\cite{shi2023mvdream} and ImageDream~\cite{wang2023imagedream} utilize multi-view diffusion models to provide highly consistent guidance in 3D object generation process. 
SyncDreamer~\cite{liu2023syncdreamer} utilizes 3D-aware attention modules to achieve synchronized multi-view image generation.
Various other works~\cite{jun2023shap, DBLP:journals/tog/ZhangTNW23, DBLP:conf/cvpr/MullerSPBKN23, DBLP:conf/siggrapha/HuiLHF22, DBLP:conf/nips/zengVWGLFK22, DBLP:journals/corr/abs-2212-08751, DBLP:journals/corr/abs-2303-17015} employ 3D data to train diffusion models for direct 3D object generation, which are fast, but struggle with output diversity.

\subsection{3D Avatar Generation}

Using strong human-body priors such as SMPL~\cite{loper2015smpl} and SMPL-X~\cite{pavlakos2019expressive}, 
it is possible to generate high-quality human avatars based on the general 3D generation methods.
EVA3D~\cite{hong2022eva3d} combines a GAN backbone with a pose-guided sampling method to generate high-quality 3D human avatars. 
AvatarCLIP~\cite{hong2022avatarclip} first solves text-to-human generation by utilizing the pre-trained CLIP model to guide optimization of geometry and color networks.

Dreamavatar~\cite{cao2023dreamavatar} and AvatarCraft~\cite{DBLP:journals/corr/abs-2303-17606} utilize SMPL to initialize the implicit human geometry used in the diffusion-guided generation process.  
DreamHuman~\cite{kolotouros2023dreamhuman} adopts ImGHum~\cite{alldieck2021imghum} as body priors and proposes a focus rendering mechanism to better reconstruct the detailed geometry of avatars.
DreamWaltz~\cite{huang2023dreamwaltz} utilizes ControlNet~\cite{zhang2023adding} to provide pose guidance to finetuning the animated representation.  
AvatarVerse~\cite{zhang2023avatarverse} and AvatarStudio~\cite{zhang2023avatarstudio} both utilize a DensePose-guided ControlNet in the generation process to circumvent the multi-face ``Janus'' problem and to support part geometry optimization.
TeCH~\cite{huang2024tech} supports image-prompt avatar generation by training an additional DreamBooth model~\cite{ruiz2023dreambooth} on the input image for the SDS guidance model. 
TADA~\cite{liao2024tada} directly distills 2D diffusion models to optimize the normals and displacements of an SMPL body mesh.

Most of these methods mainly focus on text-to-3D character generation and cannot utilize image prompts, which is necessary for controllable character generation. The DreamBooth-based methods are hampered severely by the ``Janus'' problem due to the strong front-view biases caused by overfitting on the single input image.

\section{Method}

 
\edit{We now explain the overall framework of our \method{}, which aims to efficiently generate A-pose 3D characters from 2D images in arbitrary poses. Sec.~\ref{sec:dataset} first introduces our \dataset{} dataset to show how we organize our data to assist the diffusion models in 3D spatial understanding and character pose canonicalization. Sec.~\ref{sec:2d_gen} then explains how CharacterGen generates highly consistent multi-view pose-canonicalized character images. Finally, Sec.~\ref{sec:3d_gen} shows our efficient 3D reconstruction pipeline.}

\subsection{\dataset{} Dataset}
\label{sec:dataset}

To further improve diffusion models' ability to understand 3D characters and to alleviate the ``Janus'' problem, 
\todo{we have prepared the Anime3D dataset with 13,746 stylized character subjects.} 


\subsubsection{Data Acquisition}

Existing large 3D object datasets like Objaverse~\cite{deitke2023objaverse} or OmniObject3D~\cite{wu2023omniobject3d} do not contain enough 3D stylized characters for our training purposes. 
Inspired by PAniC3D~\cite{chen2023panic}, 
we first collected a large dataset of nearly 14,500 anime characters from the VRoidHub~\cite{vroidHub} and then removed non-humanoid data, to leave 13,746 character models.

\begin{figure}[t!]
\centering
\includegraphics[width=0.5\textwidth]{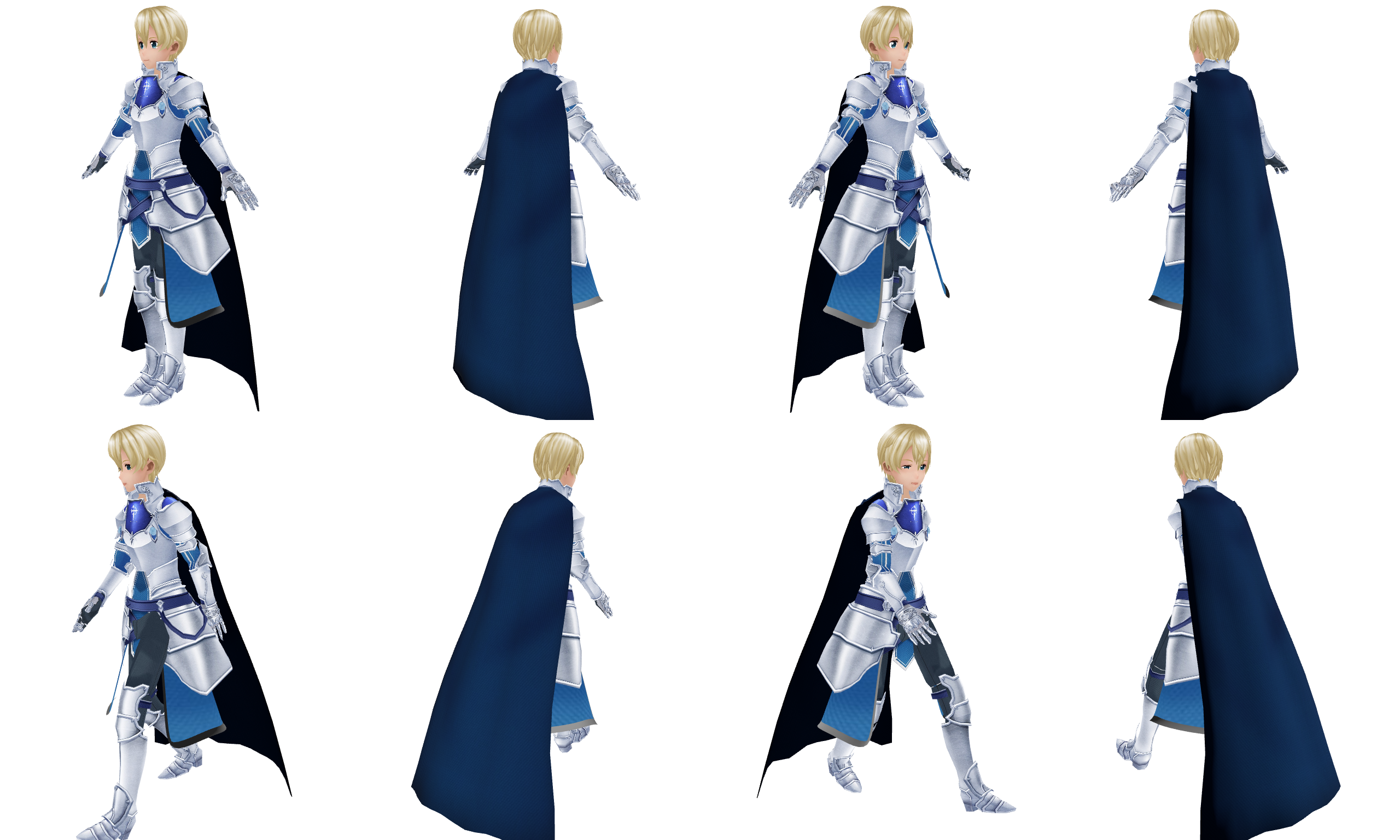}
\caption{An example character from our Anime3D dataset from four different camera views, demonstrates how we organize the image pairs during training to extend UNet's ability to determine a canonical pose.}
\label{Anime3D Dataset}
\end{figure}

\begin{figure*}[ht!]    
\centering
\includegraphics[width=1.0\textwidth]{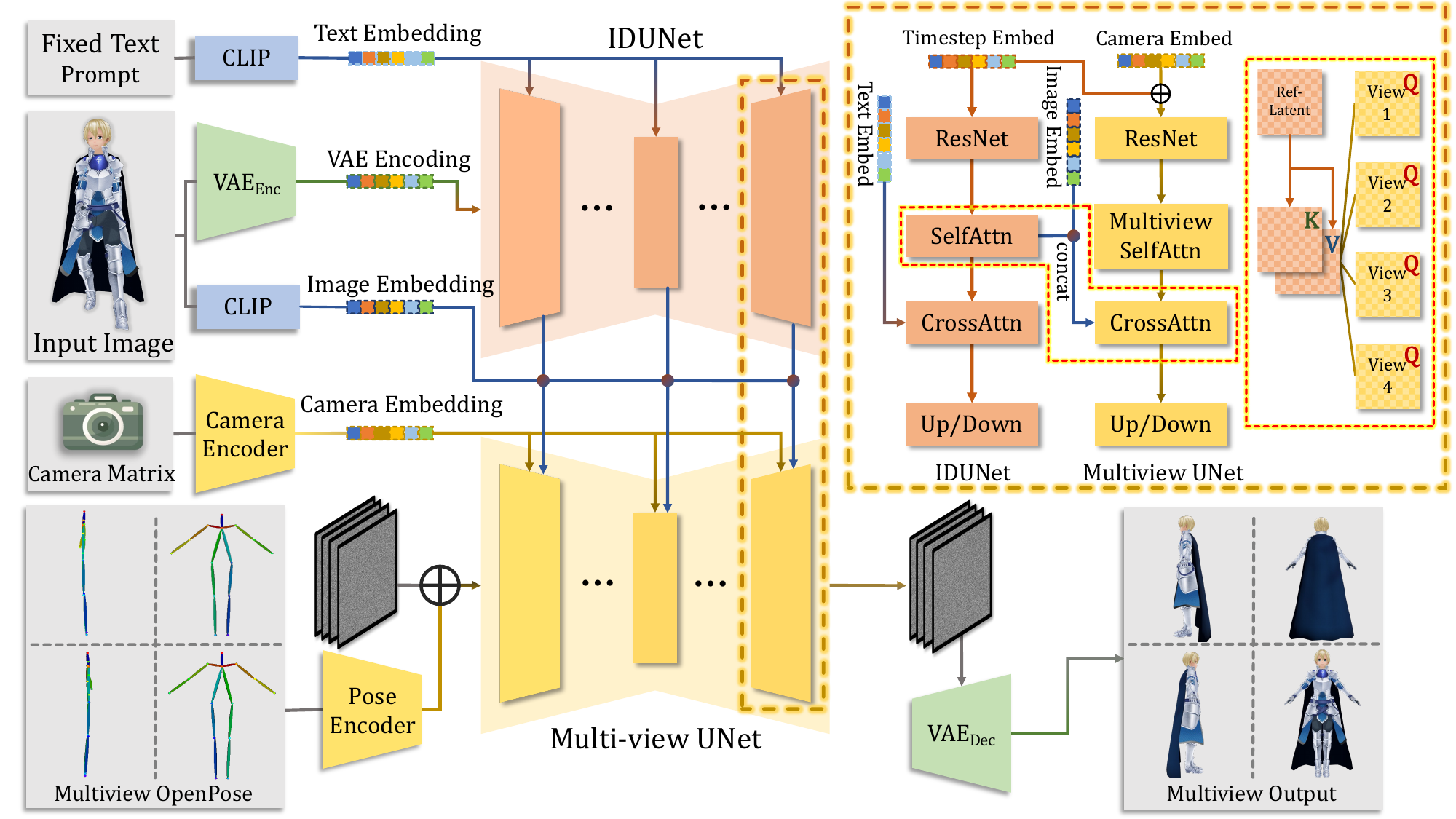}
\caption{Pipeline for generating four views of consistent images, showing how our IDUNet extracts local pixel-level features to strengthen the multi-view UNet. \final{Here "Q", "K", and "V" denote the query, key, and value matrix in the attention mechanism.}}
\label{fig:pipeline}
\end{figure*}

\subsubsection{Data Processing}

We need to render all the objects into image format to fine-tune 2D diffusion models.
We utilize the three-vrm~\cite{three-vrm} framework to render these characters.

We first obtain ``A-pose'' characters and posed characters to generate canonical pose and random pose image pairs. \final{For A-pose characters, we set the joint rotation of the left and right arm to $45^\circ$ in the $Z$-axis and the left and right upper leg to $6^\circ$ in the $Z$-axis. All other joint parameters remain untouched. For the \emph{posed} character setting, we download 10 human skeleton animations from Mixamo~\cite{mixamo}, including sitting, singing, and walking, etc. We randomly select frames from these animations and apply the corresponding motion to the VRM character models. In addition, we also randomize joints of the mouth and eyes to generate a variety of facial expressions, such as winking. We normalize the bounding boxes of the character model to $[-0.5, 0.5]^3$. We configure the camera field of view (FoV) to $40^\circ$ and set the distance between the camera and the scene origin to 1.5. The character images are rendered with ambient light and directional lighting.}

\edit{In the training process, we use four A-pose images and a single posed image as a pair because four images in orthographic views already contain sufficient appearance information for a 3D character.}
Therefore it is natural to render all objects with azimuth angles of  $\{0^\circ,90^\circ,180^\circ,$ $270^\circ\}$ and an elevation angle of $0^\circ$. To enhance the model's grasp of spatial body layout, we also render three additional groups with random initial azimuth, as depicted in Fig.~\ref{Anime3D Dataset}. We also render 4 additional views with completely random azimuth and elevation to fine-tune the \yanpei{generalizable reconstruction model (see Sec.~\ref{sec:3d_gen}).}

\subsection{\yanpei{Multi-view Image Generation and Pose Canonicalization}}
\label{sec:2d_gen}
\edit{We now consider how to generate highly consistent multi-view images from the given character image.
The overall framework is shown in Fig.~\ref{fig:pipeline}. We use our IDUNet to transfer patch-level appearance features from the input image to the multi-view denoising UNet. We also introduce a pose embedding network to provide more character layout information to assist the pose canonicalization task.
}

\subsubsection{IDUNet}

Our IDUNet aims to retain sufficient features from the original posed image and ensure high consistency between the four generated views.
\todo{Previous work, IP-Adapter~\cite{ye2023ip-adapter}, adds adapter modules into diffusion UNet structure. Appearance information in input images is transferred to the generated images via the cross-attention mechanism between the input image features and latent features. 
}
However, in practice, we observe that
IP-Adapter cannot capture the fully detailed texture from input images, as such methods only utilize the global CLIP embeddings of the condition image, which loses pixel-level details during the image encoding, leading to inconsistent results. 

To better incorporate features of the condition image,
we propose IDUNet, to introduce pixel-level guidance in the generation process. 
Inspired by ControlNet~\cite{zhang2023adding}, the structure of IDUNet is identical to the multi-view UNet. \final{IDUNet takes a fixed text-prompt "best quality" to provide general guidance in the generation process.}
%
Unlike ControlNet,
to ensure local patch-level interaction between all patches in both the denoised image and the condition image,
we leverage the cross-attention between
the latent tokens and condition image tokens
rather than merely adding them together.

Note that the IDUNet is used to provide pixel-level features into the multi-view UNet, and adding noise to the input condition image severely diminishes the texture detail of the 3D characters.
In contrast, a traditional denoising UNet is applied to a noisy image to predict noise according to the timesteps.
Thus, we directly adopt a VAE to encode the noise-free input image.

\subsubsection{Multi-view UNet}


The target of multi-view UNet is to generate multi-view A-pose images with highly consistent appearance from a single posed input image. Within the multi-view UNet, we simultaneously apply the denoising process on the four-view noisy latent $x_{\text{4v}} \in \mathrm{R}^{B \times 4 \times N \times D}$. \edit{Here $B$, $N$, and $D$ denote latent batch size, token numbers, and token feature dimension respectively.}
\final{The multi-view UNet takes the extrinsic camera matrices of the four views as spatial guidance. During the inference stage, the camera poses are set with the fixed azimuths of $\{0^\circ, 90^\circ, 180^\circ, 270^\circ\}$ and the elevation of $0^\circ$.}
\todo{The transformer block in our multi-view UNet consists of a spatial self-attention module and a cross-attention module. As in MVDream~\cite{shi2023mvdream}, the spatial self-attention module directly takes tokens of all four noisy latent $x_{\text{4v}}$ and corresponding camera view embeddings.
In the spatial self-attention layer, $x_{\text{4v}}$ is reshaped into $ (B, 4N, D) $ for patch-level cross-view interaction.
This design allows the denoising UNet to capture the global relationships across different views, ensuring image generation with high consistency. }

\begin{figure*}[ht!]
\centering
\includegraphics[width=0.95\textwidth]{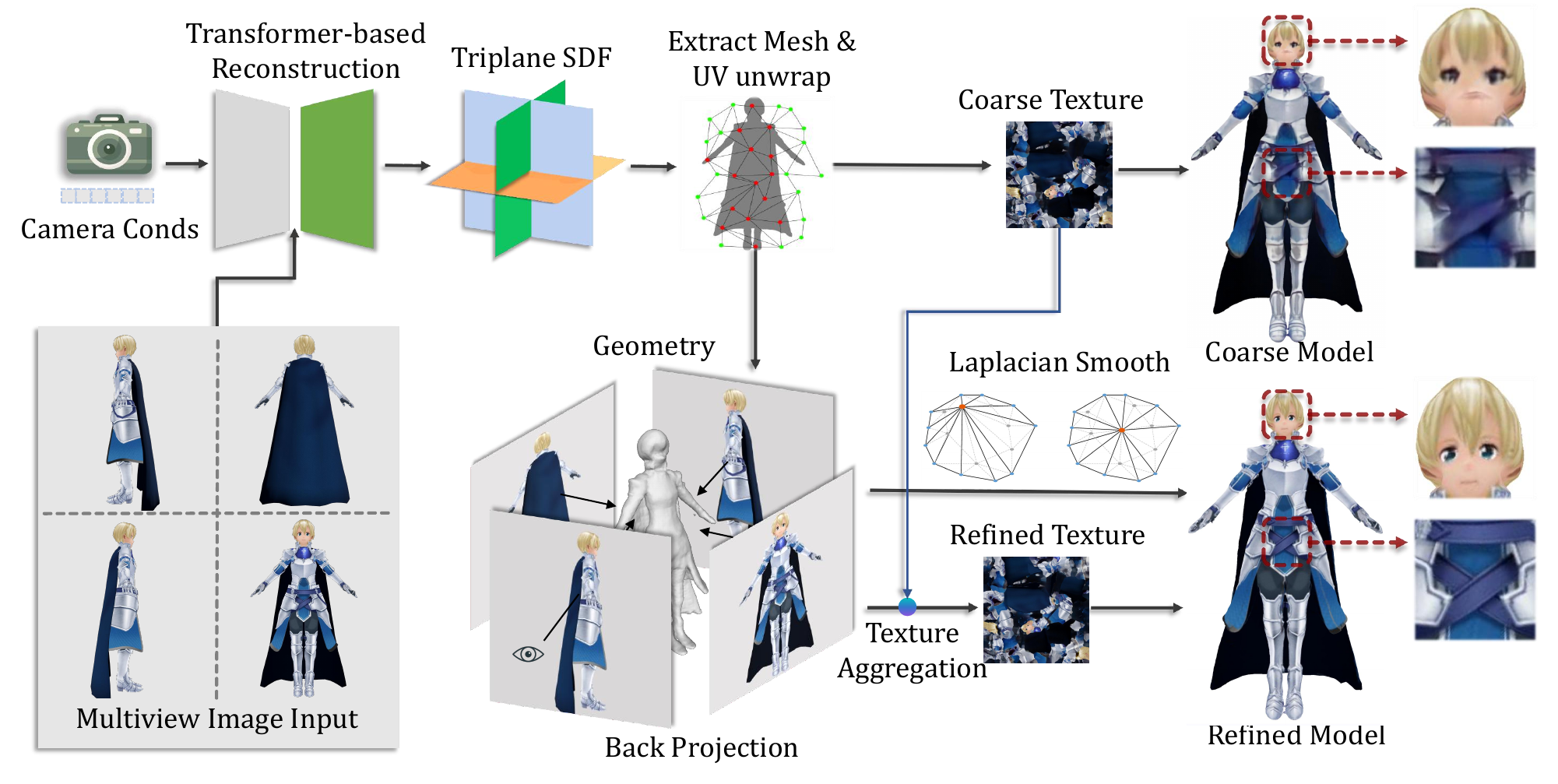}
\caption{Pipeline for generating a final refined character mesh from generated multi-view images. In the first stage, we utilize a deep transformer-based network to generate a character with a coarse texture and then use a texture back-projection strategy to enhance the appearance of the generated mesh.}  
\label{fig:lrm_fig}
\end{figure*}

In each following cross-attention module,  
\gmh{the condition features from IDUNet
$f_{\text{ID}}$} are concatenated with the CLIP-encoded image features $f_{\text{CLIP}}$ to generate final condition features $f_{\text{Cond}}$.
Then, a cross-attention layer is used to introduce patch-level interactions into $x_{\text{4v}}$. 
The whole process is given in Eq.~\ref{multi_unet} and is illustrated in Fig.~\ref{fig:pipeline}(above right).
\begin{align}
\label{multi_unet}
  f_{\text{Cond}} &= \text{concat}(f_{\text{ID}}, f_{\text{CLIP}}) \\
  x_{\text{4v}} &= \text{Cross\_Attn}(x_{\text{4v}}, f_{\text{Cond}})
\end{align}


Earlier work~\cite{lin2024common} on the diffusion process shows that
\todo{applying zero sample-noise-ratio (SNR) at the last timestamp $T$ in training improves the final generation quality
because the input noise in the inference stage is pure Gaussian noise.
To achieve zero-SNR in the training stage,
we manually set $SNR_{T}$ to zero and linearly scale other Guassian distribution parmaeters $\beta_s$. We set the UNet to directly output velocity $ v_{\text{pred}} \in \mathrm{R}^{(B, 4, N, D)} $ of the multi-view noisy latent and convert it to the noise $\epsilon_{\text{pred}}$. The final optimization target is given in Eq.~\ref{eq:noise_loss}.
}
\begin{equation}
\label{eq:noise_loss}
  L_{\text{4v}} = || \epsilon_{\text{4v}} - \epsilon_{\text{pred}} ||^2_2.
\end{equation}
where $\epsilon_{\text{4v}}$ is the noise added to the multi-view A-pose images in the diffusion forward pass. 

\subsubsection{Pose Canonicalization}


\edit{Combined with IDUNet, our Multi-view diffusion model can successfully generate highly consistent orthographic view images while maintaining detailed features from the prompt image.
To enable the diffusion model to achieve character pose canonicalization during the generation process, we jointly train the two UNets with the image pairs from our \dataset{} dataset.
However, simply training the diffusion network without extra pose constraints will lead to character layout misplacement and the emergence of unrelated body parts.}
To tackle these problems, we introduce character layouts to the diffusion models for generating A-pose character images 
by adopting OpenPose~\cite{8765346} to predict the pose embedding as an additional condition. 
The generated embeddings are directly added to the latent noise to aid the diffusion model in learning the relationships between character joints and the generated character layout.
\final{Since characters have various body shapes, we leverage three different sets of OpenPose images from our \dataset{} dataset and select the one with the highest CLIP score as the input pose condition in the inference stage.}

\subsection{3D Character Generation}
\label{sec:3d_gen}
We now describe how we efficiently 
generate 3D characters from the four-view images generated by our multi-view pose canonicalization diffusion model.
\edit{As shown in Fig.~\ref{fig:lrm_fig}, we adopt a coarse-to-fine process for the 3D character generation task.
We first utilize a two-stage transformer-based network to reconstruct the geometry and coarse appearance of the character following the design of LRM~\cite{hong2023lrm}. 
Subsequently, we employ a texture back-projection strategy to quickly improve the texture quality using the generated high-resolution four-view images. Finally, we utilize Poisson Blending~\cite{poisson} to reduce seams on the texture map. }

\subsubsection{Character Reconstruction with Coarse Texture}

\edit{Inspired by LRM~\cite{hong2023lrm}, we utilize a deep transformer network to efficiently reconstruct characters from the four-view images generated by the multi-view diffusion model. While LRM is trained using the Objaverse dataset~\cite{deitke2023objaverse} to allow versatile 3D object generation, it does not sufficiently capture the intricacies of human character layouts. To retain the reconstruction network's ability to process both general 3D objects and stylized characters, we initially pre-train our transformer network on the Objaverse dataset. Then we fine-tune the model with our \dataset{} dataset to introduce more priors of human body structure.}

\edit{Original LRM proposes to mainly train with NeRF~\cite{mildenhall2021nerf} representation. However, directly extracting geometry from NeRF models often yields noisy surface geometry, which can be problematic for the subsequent use of character meshes in downstream graphics pipelines.
Instead, we utilize a two-stage fine-tuning strategy for our reconstruction network. The first stage involves using a triplane NeRF representation similar to LRM, to establish the character's coarse geometry and appearance.
In the second stage, we modify the decoder module of our reconstruction network to predict signed distance functions (SDFs) rather than density fields, which allows CharacterGen to achieve smoother and more precise surface geometry.}

In addition to MSE loss,  
we also incorporate mask loss and LPIPS loss~\cite{zhang2018unreasonable} to supervise reconstruction appearance.
\final{We adopt the binary-cross-entropy loss between ground-truth alpha masks and rendered alpha masks as the mask loss to help the reconstruction model distinguish empty space within input four-view images.} LPIPS loss is used to extract perceptual information from input images.
The training target is given in Eq.~\ref{eq:lrm}.
\begin{align}
\label{eq:lrm}
  L_{\text{recon}} &= \lambda_1 L_{\text{mse}} + \lambda_2 L_{\text{mask}} + \lambda_3 L_{\text{LPIPS}}
\end{align}

\edit{Here, $\lambda_1$, $\lambda_2$ and $\lambda_3$ are hyperparameters, which are set to 1, 0.1 and 0.5 by default.} \final{We leverage Laplacian Smooth on the extracted meshes to further reduce the noise on the surface.}

\subsubsection{3D Character Refinement}

Our reconstruction network can rapidly reconstruct the 3D implicit representation of a character and
we can extract the final mesh along with a coarse $UV$ map from the reconstructed tri-plane with DMTet~\cite{shen2021deep}.
\edit{However, the generated mesh still lacks texture details because the DMTet-based extraction process loses appearance information during the $UV$ unwrapping process.}
To tackle this problem, we further utilize the generated four-view images to improve the quality of the generated texture maps. For efficient rasterization in this step, we employ NvDiffRast~\cite{Laine2020diffrast} as the renderer.
\final{Since the generated four-view images are in lower resolution than the texture map, multiple texels may be projected onto the same image pixel. During differentiable-rendering-based optimization, the gradients for these texels become noisy. The sparsity of input views compounds the difficulty of fixing these noisy texels, resulting in severe degradation of the output refined texture map.}
To circumvent this problem, we project the four-view images into texels in texture space and employ a depth test to remove occluded texels.
\final{We also notice that directly back-projecting the four-view images onto the mesh leads to noisy texels at the character's body silhouette. We compute the inner products of the four orthogonal camera view directions with the normal texture map. Texels with inner products greater than -0.2 are disregarded to eliminate noise around the silhouette. For texels overlapping in multiple views, we select the back-projected texels with RGB values closest to the coarse texture.}
Then we utilize Poisson Blending~\cite{poisson} to aggregate projected texels and origin texels to reduce seams in the final textures.



\section{Experiments}

\subsection{Implementation Details}

We divide our Anime3D dataset into a training set and a testing set in a 50:1 ratio. 
\final{We use the Stable Diffusion 2.1 model as the base model of both our IDUNet and multi-view UNet. The training process is carried out on 8 NVIDIA A800 GPUs for 3 days on images with $512 \times 512$ resolution and another 2 days on images with $768 \times 512$ resolution. In each training step, we jointly train both IDUNet and multi-view UNet. We begin by sampling a group of four-view images and a single-condition posed image from \dataset{}. 
Given our intention to generate four images with azimuths of $\{0^\circ, 90^\circ, 180^\circ, 270^\circ\}$, we sample this group with a probability of 0.8. We also include other four-view image groups in our training steps to strengthen the diffusion model's spatial understanding of 3D characters. We sample the front-view posed character images whose azimuth ranges from $[-90^\circ, 90^\circ]$. During the training phase, the posed image is fed into the IDUNet, and the four-view images are sent to the multi-view UNet.}

\final{To fine-tune the transformer-based reconstruction model, we first fine-tune for 50 epochs with the NeRF representation and for another 30 epochs on SDF. The fine-tuning process takes 1 day on 8 A800 GPUs.  
As \method{} does not require training in the inference steps, the entire generation pipeline (including four canonical-pose image generation, 3D character mesh reconstruction, and texture map refinement) can run on a single GPU. }



\subsection{Results and Comparison}
\label{sec:exp_main}
\edit{We conduct experiments on both 2D multi-view character image generation and 3D character mesh generation to evaluate the efficiency and effectiveness of our \method{}.}

\subsubsection{2D Multi-view Generation}

\begin{figure*}[ht!]
\centering
\includegraphics[width=0.95\textwidth]{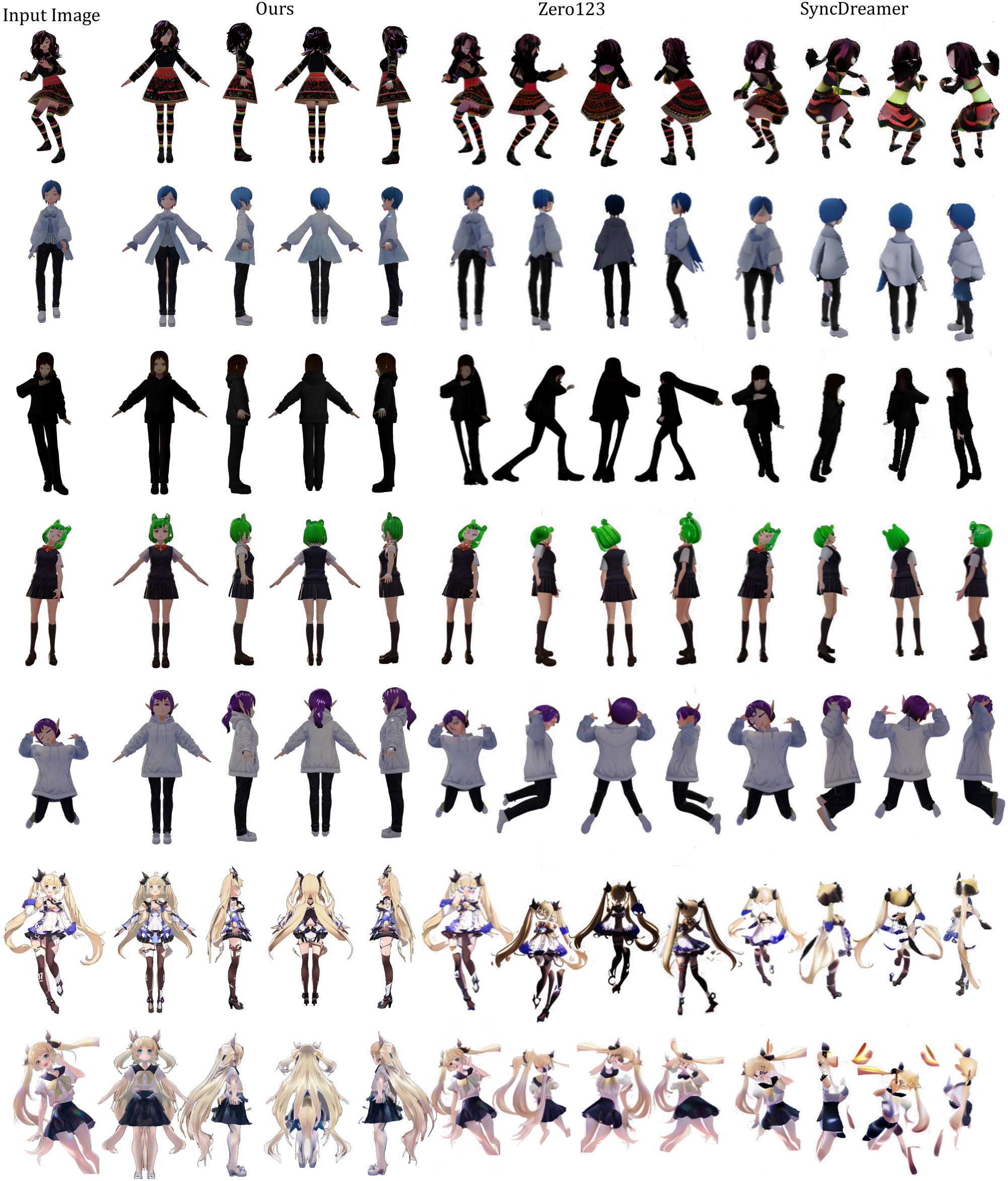}
\caption{We compare our generated four A-pose character images with other methods. The azimuths for all examples are set as $\{0^\circ, 90^\circ, 180^\circ, 270^\circ\}$. $\copyright$ kinoko7}
\label{fig:res_2d}
\end{figure*}

We test our models on images from the testing split of \dataset{} as well as online sources and compare our results to those from Zero123~\cite{liu2023zero} and SyncDreamer~\cite{liu2023syncdreamer}. The comparison results are shown in Fig.~\ref{fig:res_2d}. It can be seen that, given some difficult body poses, Zero123 and SyncDreamer struggle to preserve enough geometry and appearance information of generated images. Our \method{} adeptly performs canonical pose calibration and generates consistent character images across four views, which significantly enhances the subsequent character mesh reconstruction process.

\final{We also conduct experiments on all character images from the test split of \dataset{}. We fine-tune the baseline methods on \dataset{} for 100 epochs and show the quality metrics in the upper part of Tab.~\ref{tab:2d_metric}.}
\edit{Our calibrated A-pose images are benchmarked against ground-truth A-pose images, whereas images generated by other methods are compared with corresponding posed images. The results evaluate CharacterGen's superior generation quality and its consistency with the multi-view diffusion model.}

\begin{table}
  \setlength{\tabcolsep}{2.0mm}
  \caption{\final{We show the quantitative metrics of both 2D multi-view generation and 3D character generation methods on the test split of \dataset{} to evaluate the effectiveness of our \method{}.}}
  \label{tab:2d_metric}
  \begin{tabular}{l|c|c|c|c}
    \toprule
    Methods& SSIM$\uparrow$ & LPIPS$\downarrow$ & FID$\downarrow$ & CD$\downarrow$\\
    \midrule
    \method{}(2D) & {\bf 0.901} & {\bf 0.086} & {\bf 0.019} & - \\
    Zero123  & 0.768 & 0.224 & 1.42 & - \\ 
    Zero123(fine-tuned) & 0.813 & 0.175 & 1.34 & - \\
    SyncDreamer  & 0.807 & 0.194 & 0.396 & - \\
    SyncDreamer(fine-tuned) & 0.822 & 0.17 & 0.37 & - \\
    IP-Adapter+SDXL & 0.845 & 0.143 & 0.074 & - \\ 
    \midrule
    \method{}(3D) & {\bf 0.898} & {\bf 0.093} & {\bf 0.032} & {\bf 0.001}  \\
    Magic123 & 0.873 & 0.134 & 0.116 & 0.0034  \\ 
    ImageDream & 0.886 & 0.11 & 0.345 & 0.002 \\
  \bottomrule
\end{tabular}
\end{table}

\begin{figure*}[ht!]
\centering
\includegraphics[width=1.01\textwidth]{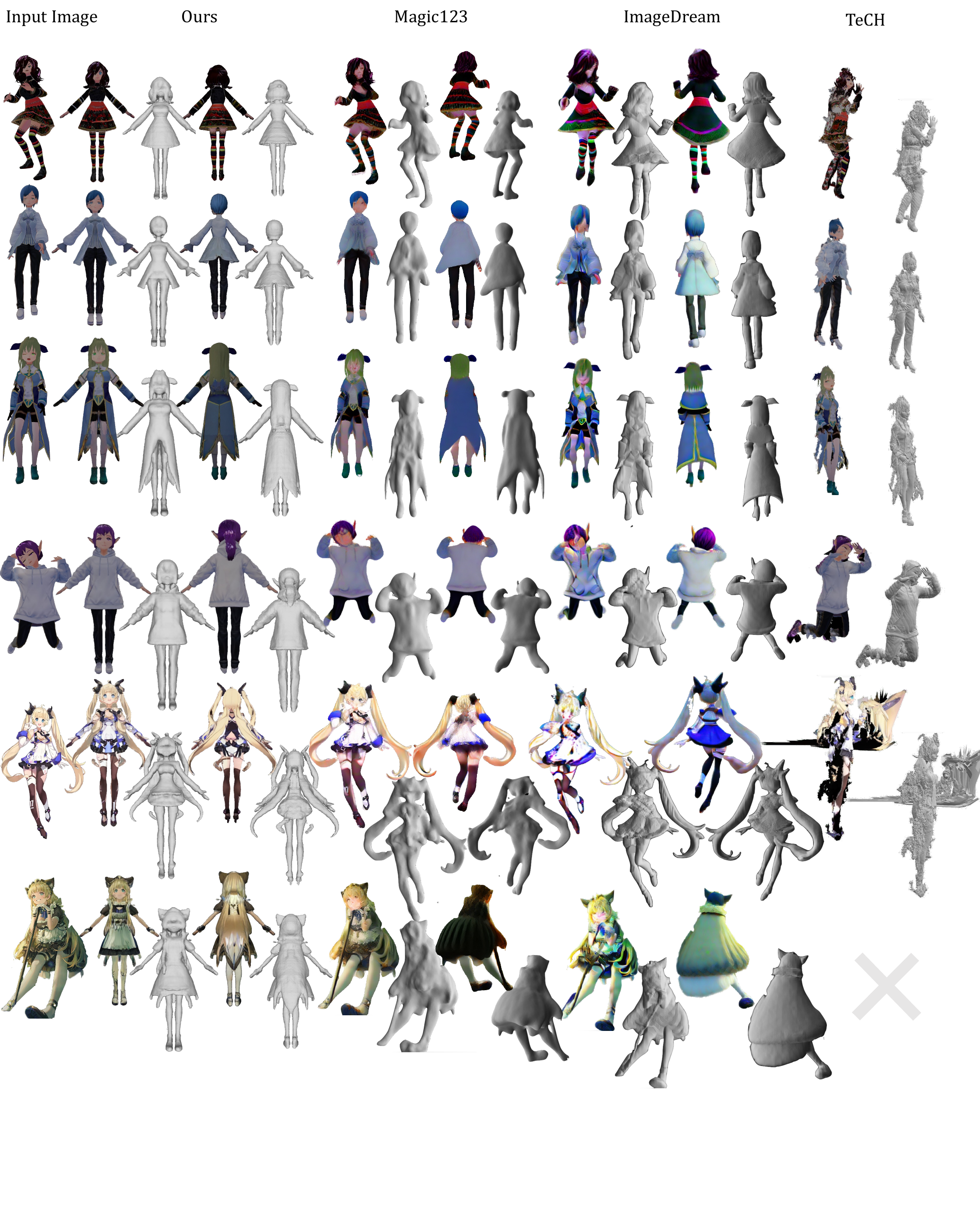}
\caption{We compare the appearance and geometry of our generated 3D characters with other methods. $\copyright$ kinoko7}
\label{fig:res_3d}
\end{figure*}

\subsubsection{3D Character Generation}

In this section, we compare our generated 3D characters with image-prompt 3D character generation methods. ImageDream~\cite{wang2023imagedream} and Magic123~\cite{qian2023magic123} all utilize the SDS-based optimization. 
TeCH~\cite{huang2024tech} extends ECON~\cite{DBLP:conf/cvpr/XiuYCTB23} and utilizes DreamBooth~\cite{ruiz2023dreambooth} to achieve image-prompt generation. We visualize the results in Fig.~\ref{fig:res_3d}. \final{We also conduct quantitative experiments on the 3D generation results on the test split of \dataset{}. The texture quality metrics are obtained by comparing rendered images and ground-truth images across the four orthogonal views. We choose Chamfer Distance(CD) as the metric to evaluate the geometry quality of generated meshes. We normalize the meshes to $[-0.5, 0.5]^3$ for calculating CD. The quantitative results are listed in the lower part of Tab.~\ref{tab:2d_metric}.}


\edit{It can be observed that our CharacterGen effectively avoids ``Janus'' problem thanks to our robust four-view reconstruction mechanism. Our generated 3D character meshes also exhibit satisfactory appearance for unseen body parts, with resourceful back-view and side-view priors from our \dataset{}. Most 3D characters generated by other methods suffer from several mesh face cohesion problems, which makes it extremely hard to rig and animate these characters. CharacterGen can successfully generate canonical pose meshes from characters with tricky poses, which facilitates downstream graphic applications. We also evaluate other methods using A-pose character images canonicalized by Animate Anyone~\cite{hu2023animateanyone}. Please refer to the supplementary materials for further details.}


\subsubsection{Comparison with IP-Adapter}

\begin{figure}[t!]
\centering
\includegraphics[width=0.45\textwidth]{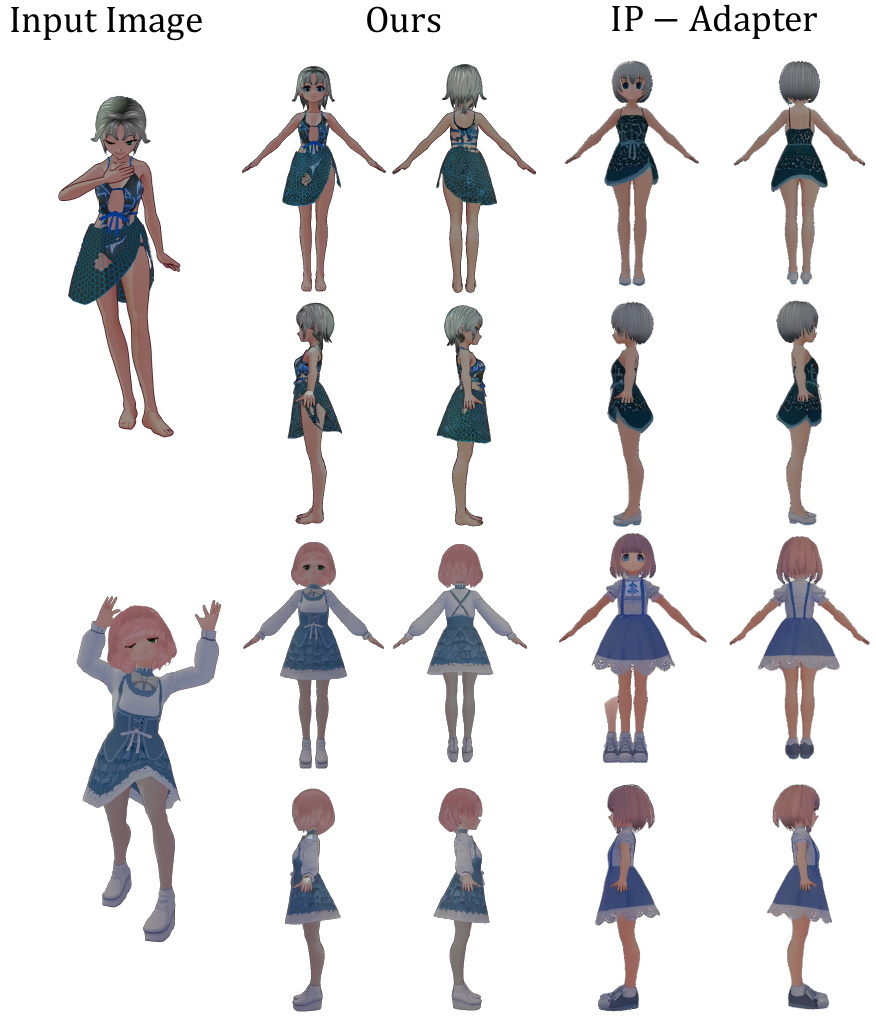}
\caption{\final{We compare the results of \method{} and IP-Adapter-SDXL.}}  
\label{fig:ipadapter}
\end{figure}

\final{Previous work IP-Adapter~\cite{ye2023ip-adapter} also supports the image-prompt generation task by incorporating adapter modules into diffusion models. We notice that IP-Adapter does not include pre-trained models for Stable Diffusion 2.1, which is the base model of our multi-view UNet. Alternatively, we train an SDXL-base model to generate $2 \times 2$ grid character images in four views with the azimuth of $\{0^\circ, 90^\circ, 180^\circ, 270^\circ\}$ for 100 epochs. The grid images are organized with a resolution of $1024 \times 1024$, which is the standard resolution of SDXL. Then we integrate official pre-trained IP-Adapter-SDXL into the base SDXL model to obtain image-conditioned multi-view results. We show the visualization results in Fig.~\ref{fig:ipadapter} and quantitative results in Tab.~\ref{tab:2d_metric}. It can be observed that IP-Adapter cannot preserve detailed appearance information and may fail to generate correct character layout, while \method{} can effectively generate high-consistent multi-view character images with our IDUNet.}

\subsubsection{Generation Speed}

\begin{table}
  \setlength{\tabcolsep}{5.0mm}
  \caption{Time to generate a single 3D character. Models loading time is excluded for all methods.}
  \label{tab:time_cost}
  \begin{tabular}{l|c}
    \toprule
    Methods&Time\\
    \midrule
    \method{} & 1min \\
    Magic123~\cite{qian2023magic123} & 70min \\
    ImageDream~\cite{wang2023imagedream} & 45min \\
    TeCH~\cite{huang2024tech} & 270min \\
  \bottomrule
\end{tabular}
\end{table}

We compare the time needed to generate a single 3D character \final{mesh} by our method and other image-prompted 3D generation methods, with the comparison results detailed in Tab.~\ref{tab:time_cost}. Our method is significantly faster than other alternatives. \final{SyncDreamer and Zero123 are used to generate multi-view images. Their time cost varies based on the chosen 3D reconstruction methods. The default NeuS~\cite{neus} reconstruction takes about 10 minutes.}

\subsection{User Study}
\label{sec:us}


\begin{table*}[t!]
  \setlength{\tabcolsep}{4.0mm}
  \caption{\final{Statistics of the user study. We display the voting results for both 2D multi-view images and 3D textured character meshes.}}
  \label{tab:us}
  \begin{tabular}{l c c c | c c}
    \toprule
    metric & \method{} & Zero123(2D) & SyncDreamer(2D) & Magic123(3D) & ImageDream(3D) \\
    \midrule
    2D multi-view style consistency & {\bf 85.4\%} & 10.5\% & 4.1\% & - & - \\
    2D multi-view consistency & {\bf 81.0\%} & 17.1\% & 1.9\% & - & - \\ 
    3D character geometry quality & {\bf 78.6\%} & - & - & 2.86\% & 18.6\% \\
    3D character texture quality & {\bf 87.1\%} & - & - & 1.9\% & 11.0\% \\
  \bottomrule
\end{tabular}
\end{table*}

\final{To better evaluate the robust generation ability of our CharacterGen, we collect 15 generated four-view character images and 10 character meshes for our user study. We compare results generated by CharacterGen with other methods mentioned in Sec.~\ref{sec:exp_main}. We ask 21 volunteers to first evaluate the style consistency between the condition images and the generated four-view images, as well as the spatial consistency within the generated multi-view images. Then, they are required to assess the geometry quality and texture quality of generated 3D character models. For each example, the volunteers are asked to vote for the one they consider to be the best. As shown in Tab~\ref{tab:us}, our CharacterGen receives a significantly higher preference compared to other methods for both 2D and 3D generation tasks.}

\begin{table}[t!]
  \setlength{\tabcolsep}{2.0mm}
  \caption{\final{We utilize CLIP score to assess style similarity between generated characters and given condition images with 2D multi-view images and 3D rendered images.}}
  \label{tab:clip_score}
  \begin{tabular}{l c | l c}
    \toprule
    method(2D) & CLIP score & method(3D) & CLIP score \\
    \midrule
    \method{} & $\mathbf{83.69}$ & \method{} & $\mathbf{79.77}$ \\
    Zero123 & 79.02 & Magic123 & 74.71 \\ 
    SyncDreamer & 75.41 & ImageDream & 73.65 \\
  \bottomrule
\end{tabular}
\end{table}

\final{We additionally quantitatively evaluate the coherence between our generated results and the input images the same as in the user study. We adopt the CLIP score as the metric and utilize ViT-B/32 as the backbone model. For 2D multi-view image generation, we calculate the CLIP scores of the input image and four generated images. For 3D character generation, we first render the 3D representation at an azimuth interval of $3^\circ$, and then compute the CLIP scores between the input image and all the rendered frames. As displayed in Tab.~\ref{tab:clip_score}, our \method{} gains superior results in both 2D and 3D generation, evaluating the robust appearance modeling ability of our IDUNet, multi-view UNet, and the subsequent reconstruction model.}

\subsection{Ablation Study}

\subsubsection{IDUNet}

\begin{figure}[t!]
\centering
\includegraphics[width=0.45\textwidth]{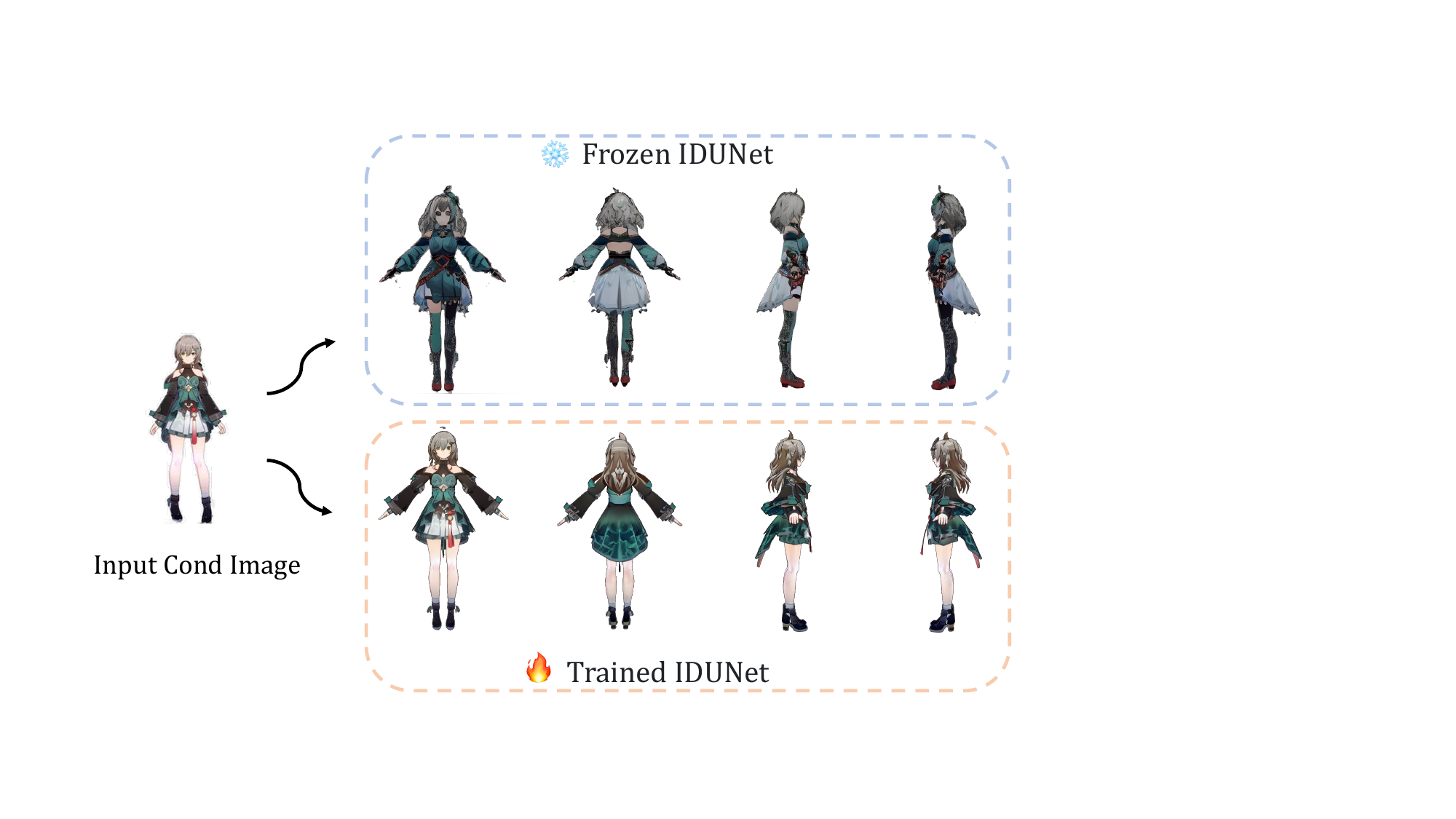}
\caption{\edit{Frozen IDUNet cannot extract sufficient appearance information from the prompt image and generates dissimilar images.}}  
\label{fig:baseline1}
\end{figure}

To demonstrate the importance of jointly training the IDUNet, we train \method{} network while freezing IDUNet with the pre-trained stable diffusion 2.1 model. The generated four-view images are shown in Fig.~\ref{fig:baseline1}. The results reveal that the generated images fail to preserve sufficient features from the input images, resulting in reduced similarity. This shows the necessity of jointly fine-tuning the IDUNet with clean posed images to enhance its ability to extract detailed clothes and facial appearance.

\subsubsection{Pose Embedding Network}

\begin{figure}[t!]
\centering
\includegraphics[width=0.45\textwidth]{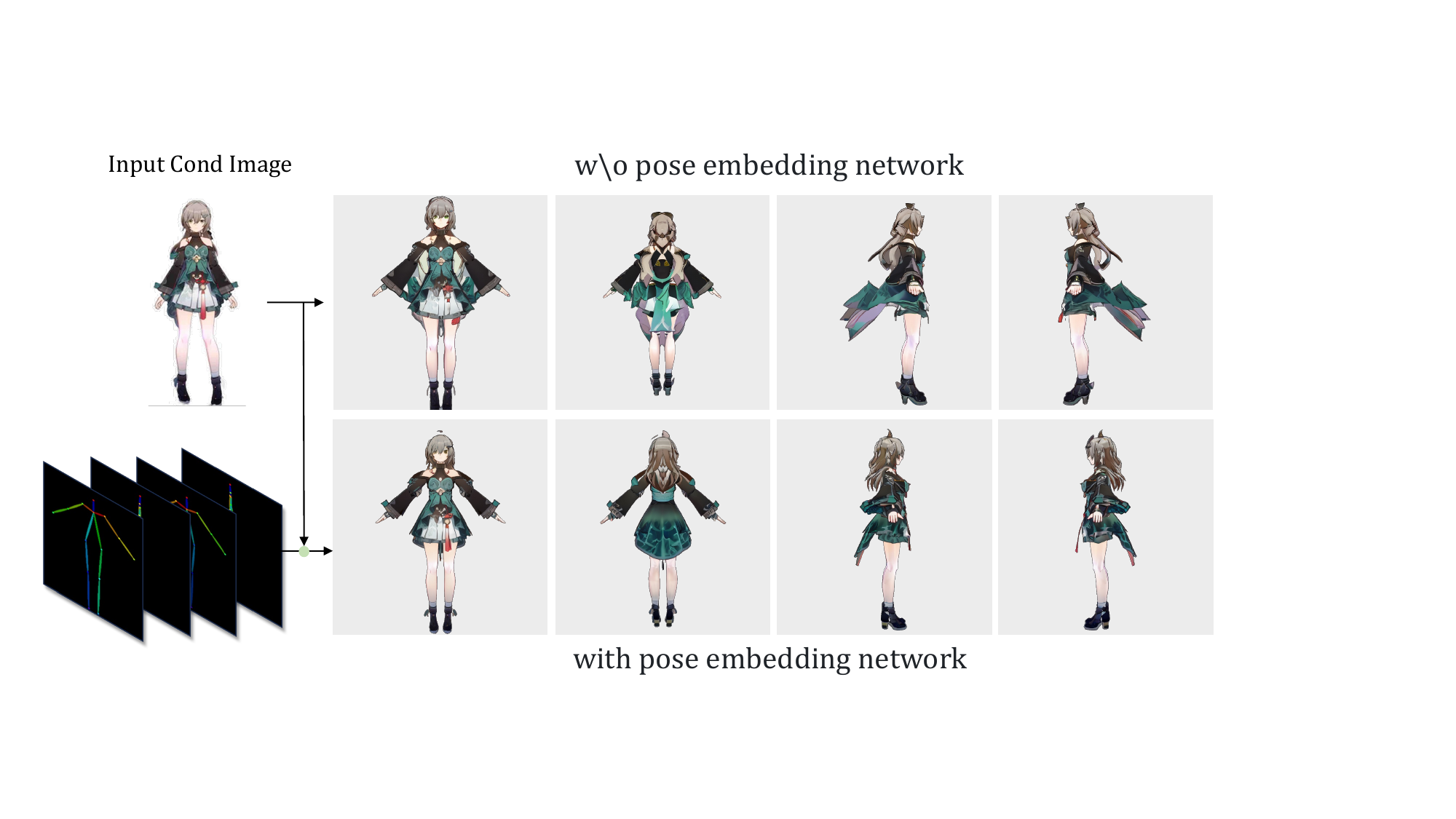}
\caption{\edit{Without the pose embedding network, the generated characters may be misplaced.}}
\label{fig:baseline2}
\end{figure}

The pose embedding network plays a crucial role in keeping character layouts in the generated four-view images. We generate additional sets of images without the pose embedding network and display representative results in Fig.~\ref{fig:baseline2}. It can be observed that the generated character images may not be located in the middle of the image in the absence of the pose embedding network. Furthermore, the lack of layout guidance can lead to the generation of inconsistent clothing parts, which could compromise 3D reconstruction in the subsequent step.

\subsection{Applications}



\begin{figure}[ht!]
\centering
\includegraphics[width=0.45\textwidth]{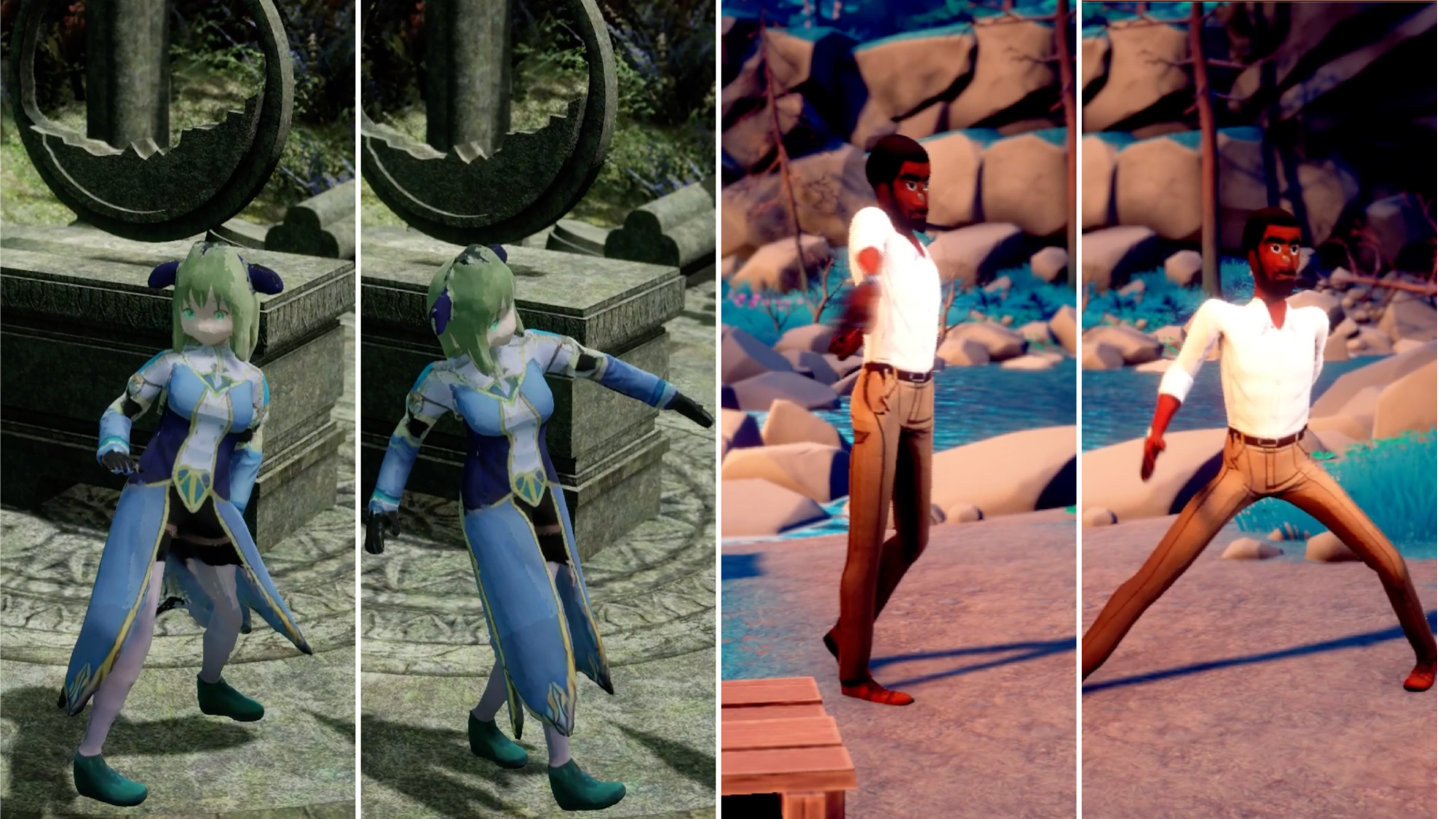}
\caption{We rig the generated characters and utilize them as 3D assets in downstream applications.}  
\label{fig:app2}
\end{figure}

\edit{CharacterGen can generate A-pose 3D characters with detailed texture maps, thereby simplifying the subsequent rigging process. We employ AccuRig~\cite{accurig} to automatically rig our generated character meshes. The rigged 3D characters can be readily utilized as animated 3D assets in various domains. 
We render various animated rigged models in Warudo~\cite{warudo} and present some results in Fig.~\ref{fig:app2}.}

\begin{figure}[t!]
\centering
\includegraphics[width=0.5\textwidth]{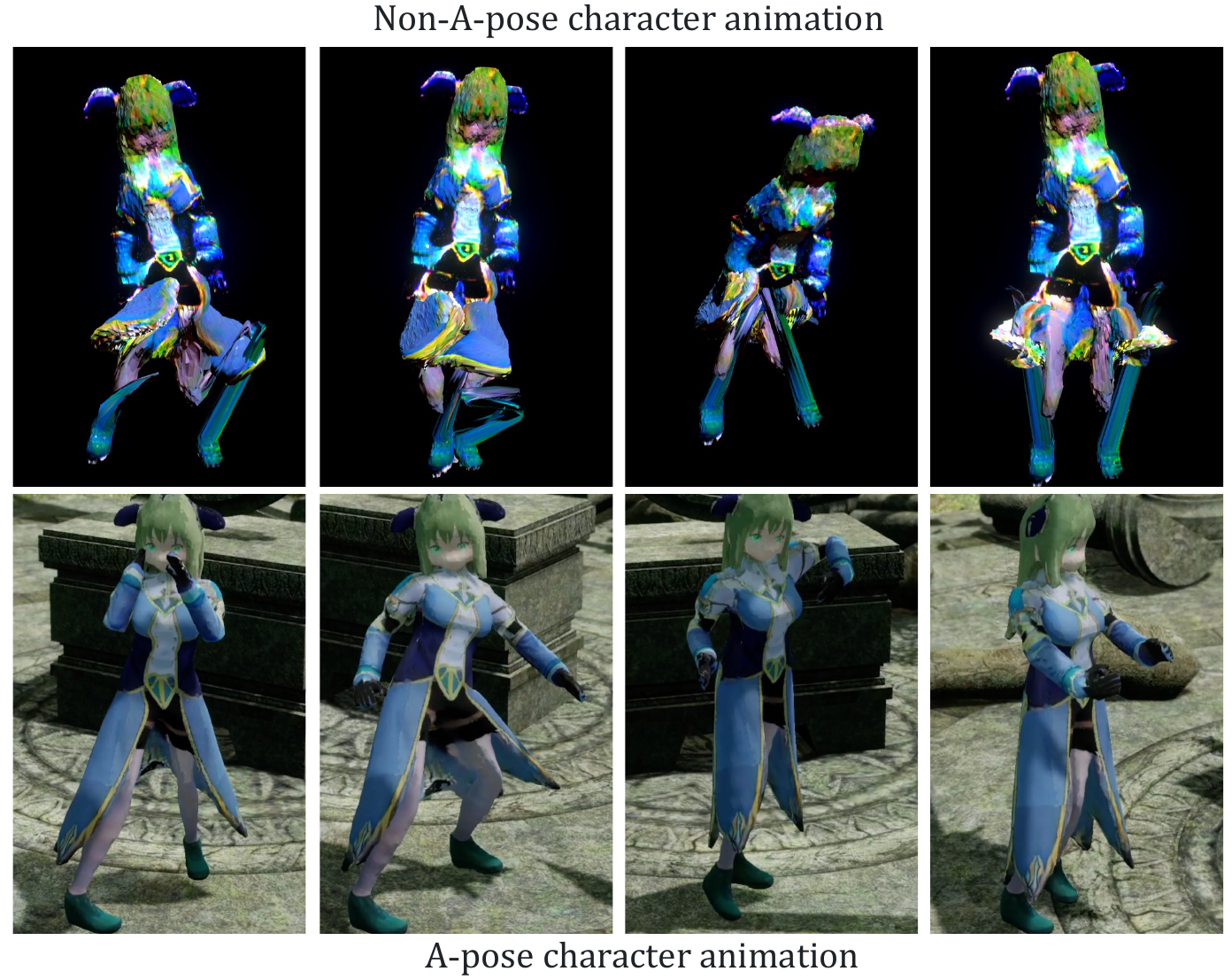}
\caption{We compare \method{}'s animated 3D characters with ImageDream~\cite{wang2023imagedream}}
\label{fig:rig}
\end{figure}

\final{To better evaluate how using an A-pose character aids skeleton rigging and animation process, we rig two 3D character meshes generated by our \method{} and ImageDream~\cite{wang2023imagedream} and visualize the animated characters in Fig.~\ref{fig:rig}. It can be observed that non-A-pose characters encounter severe mesh cohesion problem and the body structure is severely distorted while our A-posed character can be successfully animated.}

\section{Limitations and Discussion}

While our method can generate 3D characters from a single input image in an arbitrary pose, certain limitations still exist.
For the four-view A-pose image generation step, our method may not retain enough information when the character is in an extreme pose or is rendered from a non-common viewpoint. 

As for future works, the integration of additional non-photorealistic rendering (NPR) techniques into the texture refinement stage may further enhance the texture quality of the generated characters. Moreover, leveraging our trained multi-view UNet structure, it may be possible to incorporate the SDS optimization method to achieve 3D character generation with superior geometry quality.

\section{Conclusions}

This paper proposes \method{}, a novel and efficient image-prompt 3D character generation framework. We compile a new multi-pose, stylized character dataset \dataset{} to train our pipeline. Our designs include IDUNet, which extracts patch-level features from the input condition image to guide multi-view A-pose character image generation. Subsequently, we utilize a transformer-based network to reconstruct 3D character meshes and propose to utilize the texture back-projection refinement strategy to further improve the appearance of the reconstructed character meshes. Experiments demonstrate that \method{} can efficiently generate high-quality 3D characters suitable for multiple downstream applications.

\begin{acks}

\final{This work is supported by the National Science and Technology Major Project (2021ZD0112902), the National Natural Science Foundation of China (Grant No. 62220106003), and Tsinghua-Tencent Joint Laboratory for Internet Innovation Technology. The authors would like to thank Ralph R. Martin, Yuan-Chen Guo, and Yang-Guang Li for helpful discussion.}

\end{acks}

\bibliographystyle{ACM-Reference-Format}
\bibliography{sample-base}

\end{document}